\documentclass{tlp}
\usepackage{amssymb}
\usepackage{amsmath}
\newcommand{\change}[1]{#1}

\newcommand{\tuple}[1]{ (#1) }
\def\eqdef{\stackrel{\rm def}{=} \;}

\def\rXY{r_{X,Y}}
\def\cXY{c_{X,Y}}

\def\Not{\neg}
\def\qed{~\hfill$\Box$}
\def\implies{\rightarrow}
\newtheorem{define}{Definition}
\newtheorem{theorem}{Theorem}

\newtheorem{lemma}{Lemma}
\newtheorem{property}{Property}

\newtheorem{proposition}{Proposition}

\begin{document}
\bibliographystyle{acmtrans}

\submitted{10 August 2005}
\revised{5 May 2006}
\accepted{1 December 2006}
\title{Propositional Theories are Strongly Equivalent to Logic Programs}

\author[P. Cabalar and P. Ferraris]
{
Pedro Cabalar \\
Department of Computer Science, \\
University of Corunna, Spain. \\
email: \tt \change{cabalar@udc.es}
\and
Paolo Ferraris \\
Department of Computer Sciences,\\
University of Texas, Austin, USA.\\
email: \tt otto@cs.utexas.edu
}


\maketitle

\noindent
{\bf Note:} This article has been accepted for publication in
\emph{Theory and Practice of Logic Programming}, \copyright Cambridge University Press.\\
\enlargethispage{6ex}

\begin{abstract}
This paper presents a property of propositional theories under the answer sets semantics (called {\em Equilibrium Logic} for this general syntax): any theory can always be reexpressed as a strongly equivalent disjunctive logic program, possibly with negation in the head. We provide two different proofs for this result: one involving a syntactic transformation, and one that constructs a program starting from the countermodels of the theory in the intermediate logic of here-and-there.
\end{abstract}

\section{Introduction}

One of the most interesting consequences from research in declarative semantics
for Logic Programming (LP) has probably been the progressive increase in
expressiveness that the field has experienced over the years. A clear example
of this trend is the case of the {\em stable models} (or {\em answer sets})
semantics~\cite{Gel88}, which meant the first satisfactory treatment of default
negation, and whose simplicity directly allowed new extensions like the use of disjunction~\cite{Gel91} and negation~\cite{Lif92,Ino98} in the rule heads. In one of
the most general formalizations~\cite{Lif99d} of answer sets, logic programs
consist of rules whose head and body are {\em nested expressions}, that is,
propositional formulas without the connectives for implication and equivalence.
The intuitive behavior of these nested expressions is well understood: in fact, it is quite analogous to the way in which we nest operators in Prolog. To put an example of their utility, a set of rules like: 
\[
\begin{array}{rcl@{\hspace{60pt}}rcl} 
a & \leftarrow & b  
& 
a & \leftarrow & c \wedge \neg d 
\\ 
e \vee p & \leftarrow & b 
& 
e \vee p & \leftarrow & c \wedge \neg d 
\end{array} 
\] 

\noindent can be ``packed" into the single rule with nested expressions:
\begin{eqnarray} 
a \wedge (e \vee p) & \leftarrow & b \vee (c \wedge \neg d) \label{packed} \end{eqnarray}

In~\cite{Lif99d}, it was also shown that these rules with nested expressions can always be unfolded back into usual program rules (that allow disjunction and negation in the head) and, moreover, that this replacement can be done locally, regardless of the context (this last property of the transformation has the name of {\em strong equivalence}). 

In view of a commonly used abbreviation (see section 2.2), nested logic programs allow a free use of all the usual connectives for logic programs, excepting for implication, the only one that cannot be nested. This last syntactic restriction has been eventually removed thanks to the result established in~\cite{Lif01}. In that work, it was shown that the answer sets semantics is further generalized by {\em Equilibrium Logic}~\cite{Pea97}, which just deals with arbitrary propositional theories, imposing a selection criterion on their models under the intermediate logic of {\em here-and-there}~\cite{Hey30}. As a result, theories containing, for instance, a rule in the scope of a disjunction:
\begin{eqnarray}
(p \leftarrow q) \vee r \label{or-rule}
\end{eqnarray}
\noindent or a rule as body of other rule:
\begin{eqnarray}
p \leftarrow (q \leftarrow r) \label{nest-impl}
\end{eqnarray}
\noindent have now an interpretation in terms of answer sets. Although from a purely logical point of view, ruling out syntactic restrictions is a clear advantage, from the LP viewpoint one may wonder, however, what the practical interest of dealing with arbitrary formulas is. A recent example of their utility has been provided in~\cite{Fer05b}, where nonmonotone aggregates are expressed using rules as part of a body rule. To give an example, according to~\cite{Fer05b}, the following rule with an aggregate in the body\footnote{Although aggregates are out of the scope of this note, the intuitive meaning of this rule would be to make $p$ true whenever the addition of the numeric weights in (\ref{aggrrule}) for those atoms in $\{q,r\}$ that are true is greater than or equal to zero.}:
\begin{eqnarray}\label{aggrrule}
p & \leftarrow & 0 \leq \{q=1,r=-1\}
\end{eqnarray}
\noindent is translated as~(\ref{nest-impl}).

Despite of their interest, the intuition on how arbitrary theories may behave as ``programs'' was still far from obvious. Until now, we missed a formal connection with the usual rule format of pairs of head and body. Furthermore, we did not even know whether they provide a real increase in expressiveness, or perhaps they can always be translated instead into a strongly equivalent program (as in the case of nested expressions). In this paper we show that, in fact, the latter happens to be the case. In other words, the main result states that:
\begin{quote}
every propositional theory is strongly equivalent to a logic program.
\end{quote}

\noindent From an LP reading, this result is pointing out that any generic propositional theory can be seen as ``shorthand'' for a logic program.  Furthermore, the transformation from a propositional theory to a logic program is modular, i.e., it can be done formula by formula. For instance, formula~(\ref{nest-impl}) inside any theory can be simply replaced by the rules:
\begin{eqnarray*}
p & \leftarrow & \neg r \\
p & \leftarrow &  q \\
p \vee \neg q \vee r & \leftarrow & 
\end{eqnarray*}

\noindent and similarly (\ref{or-rule}) can be replaced by:
\begin{eqnarray*}
p \vee r & \leftarrow & q \\
\neg q \vee r & \leftarrow & \neg p
\end{eqnarray*}

Another possible reading of the main result is that the form of logic programs is a kind of {\em normal form} for arbitrary formulas from the point of view of Equilibrium Logic.

The main result is independently proven in two ways: (1) with a syntactic
recursive transformation, and (2) with a method that builds a strongly
equivalent program starting from the countermodels (in the logic of
here-and-there) of the propositional theory. Presenting the two proofs is interesting for several reasons. First, we get in this way an interesting analogy with the two ways of transforming a formula into CNF in classical logic: using a syntatic transformation or building the formula from its countermodels. Second, both proofs may provide technical tools for achieving other interesting results, as we explain later in the conclusions.

The rest of the paper is organized as follows. Section~\ref{sec:prelim} contains the preliminaries, with brief reviews of the logic of here-and-there and of answer sets semantics. In Section~\ref{sec:recurs} we present the first proof of the main result, based on the recursive syntactic transformation, whereas Section~\ref{sec:nonnested} contains the second proof, which builds a program from the here-and-there countermodels of the original theory. In Section~\ref{sec:nf} we further explore the analogies to classical logic, and obtain a disjunctive normal form based on the here-and-there models.

\section{Preliminaries}
\label{sec:prelim}

Before introducing the logic of here-and-there, a small remark about notation. We assume that we handle a set of atoms $\Sigma$ called the {\em propositional signature}. We will use letters $X, Y, \dots$ to denote propositional interpretations in classical logic, represented here as sets of atoms, that is, subsets of $\Sigma$. For any propositional formula $F$, we adopt the usual notation $X \models F$ to stand for $X$ {\em satisfies} $F$ under classical logic.

\subsection{Logic of here-and-there}

The logic of here-and-there was originally defined in~\cite{Hey30}. A {\em propositional formula} is any combination of atoms
in $\Sigma$ with connectives $\bot$ (false), $\vee$, $\wedge$, and $\implies$. We also define the abbreviations:
\begin{eqnarray*}
\neg F & \eqdef & F \implies \bot \\
\top   & \eqdef & \bot \implies \bot \\
F \equiv G & \eqdef & (F \implies G) \wedge (G \implies F)
\end{eqnarray*}
\noindent As usual, a {\em propositional theory} is a (possibly infinite) set of
propositional formulas.

The semantics of the logic of here-and-there is defined as follows. An {\em
interpretation} is a pair $\tuple{X,Y}$ of sets of atoms (respectively called
``here'' and ``there'') such that $X \subseteq Y$. Intuitively, atoms in $X$
are considered to be true, atoms not in $Y$ are considered to be false, and the
rest ($Y-X$) are thought to be undefined. We say that an interpretation $(X,Y)$
is {\em total} when $X=Y$ (no undefined atoms).

\begin{define}[satisfaction of formulas]
\label{def:satisf}
We recursively define\footnote{We have slightly simplified the definition in comparison with the usual definition of satisfaction in the logic of here-and-there which is typically provided in terms of a Kripke structure as in intuitionistic logic, but just handling two worlds in this case. It can be easily seen that both definitions are equivalent.} when an interpretation $(X,Y)$ {\em satisfies} a formula $F$, written $(X,Y)\models F$, as follows:
\begin{itemize}
\item
for any atom $a$, $(X,Y)\models a$ if $a\in X$,
\item
$(X,Y)\not \models \bot$,
\item
$(X,Y)\models F\wedge G$ if $(X,Y)\models F$ and $(X,Y)\models G$,
\item
$(X,Y)\models F\vee G$ if $(X,Y)\models F$ or $(X,Y)\models G$,
\item
$(X,Y)\models F\implies G$ if $(X,Y) \models F$ implies $(X,Y)\models G$, and $Y\models F\implies G$.
\end{itemize}
\qed
\end{define}

Although we use the same symbol `$\models$' for satisfaction in classical logic
and the logic of here-and-there, ambiguity is avoided in view of the difference in the form of the interpretation on the left. In this way, the last line of
Definition~\ref{def:satisf} is referring to classical satisfaction for $Y
\models F\implies G$. As usual, an interpretation is a {\em model} of a theory
$T$ if it satisfies all the formulas in $T$. Two formulas (theories) are {\em
equivalent} if they have the same models.

As a first immediate observation about Definition~\ref{def:satisf}, note that when the interpretation is total ($X$=$Y$), $(Y,Y) \models F$ simply collapses into classical satisfaction $Y \models F$. Another interesting property we will use later is that truth in the ``here'' component implies truth in the ``there'' component:
\begin{property}
\label{prop1}
For any interpretation $\tuple{X,Y}$ and any propositional theory $T$: \\
if $(X,Y) \models T$ then $(Y,Y) \models T$ (i.e., $Y \models T$).
\end{property}

When dealing with countermodels, this property can be rephrased as: if $(Y,Y)$ is a total countermodel of $T$, then any $(X,Y)$ is also a countermodel of $T$. Besides, using the definition of $\neg F$ as $F \implies \bot$, it also allows us to obtain the following characterization of satisfaction for negated formulas:
\begin{property}
\label{prop2}
For any interpretation $\tuple{X,Y}$ and any formula $F$: \\
$(X,Y) \models \neg F$ iff $Y \models \neg F$.
\end{property}

Axiomatically, the logic of here-and-there is intermediate between
intuitionistic and classical logic. Recall that a natural deduction system for
intuitionistic logic can be obtained from the corresponding classical
system~\cite[Table~3]{bib93} by dropping the law of the excluded middle
$$F\vee\neg F$$ from the list of postulates. The logic of here-and-there, on
the other hand, is the result of replacing the excluded middle in the classical
system with the weaker axiom schema~\cite{DeJ03}:
\begin{equation}
\label{weakmiddle}
F\vee(F\implies G)\vee\neg G. 
\end{equation} 

In addition to all intuitionistically provable formulas, the set of theorems of
the logic of here-and-there includes, for instance, the weak law of the
excluded middle 
$$ 
\neg F\vee\neg\neg F 
$$ 
and De Morgan's law 
$$ 
\neg(F\wedge G)\equiv \neg F \vee \neg G 
$$ 
(the dual law can be proved even intuitionistically).

The logic of here-and-there differs from intuitionistic logic also as far as minimal adequate sets of connectives are concerned. In fact, a disjunction
$$
F\vee G
$$
is equivalent~\cite{Luk38}, in the logic of here-and-there, to
$$
((F\implies G)\implies G)\wedge ((G\implies F)\implies F).
$$

\subsection{Logic programs}
\label{sec:LP}
The set of logic programs can be defined as a subset of propositional
formulas as follows. A {\em nested expression} is any propositional formula not
containing implications of the form $F \implies G$ with $G \neq \bot$ (i.e.,
negations and $\top$ are allowed). A {\em rule} $r$ is a formula of the form $F
\implies G$ where $F$ and $G$ are nested expressions\footnote{Traditionally,
rules $F \implies G$ are written in the form $G \leftarrow F$ we used in the introduction. Besides, $\neg$ is usually written as $\hbox{\rm
\em not}$, $\wedge$ as comma and $\vee$ as semicolon.}. The {\em head} and the
{\em body} of $r$ are respectively defined as $head(r)=G$ and $body(r)=F$. When
convenient, we will implicitly consider any nested expression $G$ as the rule
$\top \implies G$. 

As usual, a {\em literal} is any atom $a$ or its negation $\neg a$. A rule is said to be {\em nonnested} when it has the form:
$$ 
(l_1\wedge\dots\wedge l_m) \implies (l_{m+1}\vee\dots\vee l_n) 
$$ 
\noindent $(0\leq m\leq n)$ where $l_1,\dots,l_n$ are literals. An empty conjunction ($m=0$) is understood as the formula $\top$ and analogously, an empty disjunction ($n=m$) is represented as $\bot$. 

A {\em (logic) program} is a propositional theory consisting of rules; 
a program is {\em nonnested} when all its rules are nonnested.

The usual definition of the {\em answer sets} ({\em stable models}) of a logic program $\Pi$ is given in terms of the minimal models of a reduct program $\Pi^X$ constructed from $\Pi$ and a given interpretation $X$. The original answer sets semantics was defined in~\cite{Gel88} and~\cite{Gel91} for special cases of nonnested programs, and extended in~\cite{Lif99d} for programs with nested expressions. However, these definitions were later subsumed by a general logical encoding called {\em Equilibrium Logic}~\cite{Pea97}, which allows considering answer sets (or {\em equilibrium models}) for arbitrary propositional theories as a particular type of selected models in the logic of here-and-there:
 
\begin{define}[Answer set/Equilibrium model]
A set $Y$ of atoms is an {\em answer set} (or {\em equilibrium model}) for a theory $T$, if for every subset $X$ of $Y$, $(X,Y)\models T$ iff $X=Y$. 
\end{define}

An alternative, reduct-based description of equilibrium models of arbitrary propositional theory was recently obtained in~\cite{Fer05b}.

Apart from constituting a monotonic framework in which answer sets can be defined, the logic of here-and-there further satisfies an interesting property: it allows capturing the concept of strong equivalence of theories. Two theories $T_1$ and $T_2$ are {\sl strongly equivalent} if, for any theory $T$, $T_1\cup T$ and $T_2\cup T$ have the same answer sets. The main result in~\cite{Lif01} asserts that this condition holds iff $T_1$ and $T_2$ are equivalent in the logic of here-and-there.

Our main result can be enunciated now as the following theorem:
\begin{theorem}[Main result]\label{th:gen}
Every propositional theory (of Equilibrium Logic) is strongly equivalent to a logic program.\qed
\end{theorem}

\section{Proving Theorem~\ref{th:gen} by syntactic transformation}
\label{sec:recurs}

In this proof of Theorem~\ref{th:gen}, we identify every finite program with the conjunction of its rules. We also need a few lemmas.

\begin{lemma}\label{specialeq}
For any formulas $F,G$ and $K$,
\begin{equation}\label{f1}
(F\implies G)\implies K
\end{equation}
is equivalent to
\begin{equation}\label{f2}
\begin{split}
&(G\vee \neg F)\implies K\\
&K \vee F\vee \neg G.
\end{split}
\end{equation}
\noindent in the logic of here-and-there.
\end{lemma}

\begin{proof} 
In the proof of the implication from~(\ref{f2}) to~(\ref{f1}),
we assume $F\implies G$, and we want to prove $K$. We consider the three cases
$K$, $F$ and $\neg G$ from the last formula of~(\ref{f2}). The first case is
trivial. If $F$ then $G$, and if $\neg G$ then $\neg F$ from the hypothesis
$F\implies G$. In both cases, the antecedent of the implication in~(\ref{f2}) is
true, so we can conclude $K$ as well.

Now we assume~(\ref{f1}) and we want to derive each formula of~(\ref{f2}). For the first
one, we also assume $G\vee \neg F$, and we want to derive $K$. It is sufficient to notice
that both $G$ and $\neg F$ make the antecedent of~(\ref{f1}) --- and consequently $K$ ---
true. To prove the second formula of~(\ref{f2}), we consider the axiom~(\ref{weakmiddle}).
The claim is clearly proven in the cases $F$ and $\neg G$. In the remaining case, we
obtain $K$ by modus ponens on $F\implies G$ and~(\ref{f1}).
\end{proof}

\begin{lemma}\label{lemma:gen}
The implication of two finite programs is equivalent to a finite program.
\end{lemma}

\begin{proof} 
We shall prove that the implication of two programs
$\Pi_1\implies \Pi_2$
is equivalent to a program, by strong induction on the number of rules of
$\Pi_1$. If $\Pi_1$ is empty (i.e., it is $\top$) then $\Pi_1\implies \Pi_2$
is clearly equivalent to $\Pi_2$. If $\Pi_1$ consists of a single rule
$F\implies G$ then
\begin{equation*}
\begin{split}
\Pi_1\implies \Pi_2&
=
(F\implies G) \implies 
\big(\bigwedge_{(H\implies K)\in \Pi_2} (H\implies K)\big)\\
&\Leftrightarrow
\bigwedge_{(H\implies K)\in \Pi_2}
((F\implies G) \implies (H\implies K)).\\
\end{split}
\end{equation*}
It remains to notice that each conjunctive term
$(F\implies G) \implies (H\implies K)$ is equivalent to the
conjunction of two rules. 
Indeed, by Lemma~\ref{specialeq},
\begin{equation*}
\begin{split}
(F\implies G) \implies (H\implies K)
&\Leftrightarrow
H\implies ((F\implies G)\implies K)\\
&\Leftrightarrow
H\implies 
(((G\vee \neg F)\implies K)\wedge (K \vee F\vee \neg G))\\
&\Leftrightarrow
(H\implies ((G\vee \neg F)\implies K))
\wedge (H\implies (K \vee F\vee \neg G))\\
&\Leftrightarrow
((H\wedge(G\vee \neg F))\implies K)
\wedge (H\implies (K \vee F\vee \neg G))\\
\end{split}
\end{equation*}

It remains to consider the case when $\Pi_1$ contains more than one rule. In this case,
$\Pi_1$ can be broken into two programs $\Pi'_1$ and $\Pi'_2$ with
$|\Pi'_1|\leq |\Pi''_1|<|\Pi_1|$. Then, since
$$
\Pi_1\implies \Pi_2
\Leftrightarrow
(\Pi'_1\wedge \Pi''_1)\implies \Pi_2
\Leftrightarrow
\Pi'_1\implies (\Pi''_1\implies \Pi_2)
$$
the assertion follows by applying the induction hypothesis twice.
\end{proof}

\begin{proof}[Proof of Theorem~\ref{th:gen}.]

\noindent It is sufficient to prove that any theory consisting of a single formula is equivalent to a
finite program, by structural induction. We also assume that such a formula contains atoms

and connectives $\bot$, $\implies$ and $\wedge$ only (all the other connectives can be
eliminated). Clearly, every atom and the connective $\bot$ are programs. For formulas of
the form $F\wedge G$ and $F\implies G$, we assume, as induction hypothesis, that there are
two finite programs equivalent to $F$ and $G$ respectively. Clearly, $F\wedge G$ is
equivalent to the union of such two programs. The existence of a finite program equivalent
to $F\implies G$ follows from the induction hypothesis and Lemma~\ref{lemma:gen}.
\end{proof}

To understand how an arbitrary propositional theory is converted into a 
logic program, consider formula~(\ref{or-rule}) as a logic program. First
of all, we need to remove the disjunction, so we write it as
\begin{equation}
((r\implies (q \implies p)) \implies (q \implies p)) \wedge 
(((q \implies p) \implies r) \implies r).
\label{ex1}
\end{equation}
Then we convert each subformula that is not a logic program into a logic
program bottom-up.
Lemma~\ref{lemma:gen} shows how each subformula can be
converted, but simplifications or other transformations can also be applied.
For instance, by Lemma~\ref{specialeq},
the subformula $r\implies (q \implies p)$ is converted into
$$
((q\wedge(r\vee \neg \bot))\implies p) \wedge
(q\implies (p \vee \bot\vee \neg r))
$$
which can be equivalently rewritten as
$$
(q\wedge r)\implies p.
$$
Consequently, the first conjunctive term of~(\ref{ex1})
becomes
$$
((q\wedge r)\implies p) \implies (q \implies p);
$$
this formula, by Lemma~\ref{lemma:gen} again, can be rewritten as
$$
((q\wedge(p\vee \neg (q\wedge r)))\implies p)
\wedge (q\implies (p \vee (q\wedge r)\vee \neg p))
$$
which can be simplified into
$$
((q\wedge \neg r)\implies p)
\wedge (q\implies (p \vee r\vee \neg p)).
$$

With similar steps, we can rewrite the second conjunctive term of~(\ref{ex1})
as a logic program. As~(\ref{ex1}) becomes the conjunction of two programs,
it becames a program itself.

\section{Proving Theorem~\ref{th:gen} by using countermodels}
\label{sec:nonnested}

The main idea of this technique is to start from the countermodels (under the logic of here-and-there) of some propositional theory $T$ with a finite signature and construct a logic program, call it $\Pi(T)$, which has exactly the same set of countermodels. We can think about the construction of $\Pi(T)$ as a process where we start with an empty program, which would have as models all the possible interpretations, and go adding a rule per each model we want to remove. Note that, since there can only be a finite number of models and countermodels for a finite signature, $\Pi(T)$ is finite even if $T$ is infinite. In the case of an infinite signature for $T$, we can define $\Pi(T)$ as the union of $\Pi(\{F\})$ (computed on the finite set of atoms that occur in $F$) for each $F\in T$. The form of each rule in $\Pi(T)$ is defined as follows.

\begin{define}[$\rXY$]
Given an interpretation $\tuple{X,Y}$ under some propositional signature $\Sigma$, we define $\rXY$ as the (nonnested) rule:
\begin{eqnarray*}
\big( \bigwedge_{b \in X} b \big) \ \wedge \ \big( \bigwedge_{c \in \Sigma - Y} \Not c \big)  & \implies & \bigvee_{a \in Y-X} (a \vee \Not a)
\end{eqnarray*}
\qed
\end{define}
For instance, if $\Sigma=\{p,q,r\}$ then $r_{\{q\}, \{p,q\}}$ is
\begin{eqnarray*}
q\wedge \neg r &\implies &p\vee\neg p.
\end{eqnarray*}
When the interpretation is total, the head $\tuple{Y,Y}$ of $r_{Y,Y}$ is empty, leading to the constraint:
\begin{eqnarray*}
\big( \bigwedge_{b \in Y} b \big) \ \wedge \ \big( \bigwedge_{c \in \Sigma - Y} \Not c \big) & \implies & \bot 
\end{eqnarray*}
For instance, with $\Sigma=\{p,q,r\}$, $r_{\{q\}, \{q\}}$ is
\begin{eqnarray*}
q\wedge \neg p\wedge \neg r &\implies &\bot.
\end{eqnarray*}

From the definition of satisfaction in the logic of here-and-there, it is not difficult to see that $(X,Y)$ is always a countermodel of $\rXY$. Moreover, when $X \subset Y$, this is indeed the {\em only} countermodel. When $X=Y$, however, Property~\ref{prop1} may lead to additional countermodels, as reflected by the following proposition:

\begin{proposition}
\label{prop:cm}
Given any interpretation $(X,Y)$, an interpretation $(X',Y')$
is a countermodel of $\rXY$ iff
\begin{enumerate}
\item[i)] $Y'=Y$, \ \ \ \ \ \ \ \ \ \ \; if $X=Y$, and
\item[ii)] $X'=X$, $Y'=Y$ \, otherwise.
\end{enumerate}
\end{proposition}

To prove this result, the following lemma will be particularly useful:

\begin{lemma}
\label{lem:bodyrI}
Let $(X,Y)$ and $(X',Y')$ be a pair of interpretations. Then,
$$(X',Y') \models body(\rXY)\text{ iff }
X \subseteq X' \subseteq Y' \subseteq Y.$$
\end{lemma}

\begin{proof} Note first that $X' \subseteq Y'$ trivially follows from the
form of interpretations. We have that
$(X',Y')\models body(\rXY)$ iff
$$
\text{$(X',Y') \models b$ for all $b \in X$, and
$(X',Y') \models \neg c$ for all $c \in \Sigma-Y$.}
$$
By Property~\ref{prop2}, we can rewrite this condition as
$$
\text{$b\in X'$ for all $b \in X$, and
$c\not\in Y'$ for all $c \not\in Y$}
$$
and then as
$$
\text{$b \in X\Rightarrow b\in X'$, and $c \not\in Y\Rightarrow c\not\in Y'$}.
$$
This is clearly equivalent to $X\subseteq X'$ and $Y'\subseteq Y$.
\end{proof}

\medskip
\begin{proof}[Proof of Proposition~\ref{prop:cm}]
\begin{enumerate}
\item[i)] If $X=Y$ then $head(\rXY)=\bot$ so that $\rXY$ can be rewritten as $\neg body(\rXY)$. Consequently, by Property~\ref{prop2}, the countermodels of $\rXY$ are the interpretations $(X',Y')$ such that $Y'$ satisfies $body(\rXY)$ in classical logic. It remains to notice that, in this case, $body(\rXY)$ is the conjunction of all literals classically satisfied by $Y$, and so, this guarantees $Y'=Y$.

\item[ii)] If $X \neq Y$, we must have $X \subset Y$, by the form of interpretations. We show first that $(X,Y)$ is not model of $\rXY$. To this aim, it suffices to show that $(X,Y) \models body(\rXY)$ but $(X,Y) \not\models head(\rXY)$. The satisfaction of the body trivially follows from Lemma~\ref{lem:bodyrI} (take $X'=X$ and $Y'=Y$). For the head, take the disjunction $a \vee \neg a$ in $head(\rXY)$ for each $a \in Y-X$. Since $a \not\in X$ we get $(X,Y) \not\models a$ whereas since $a \in Y$, $Y \not\models \neg a$ and, by Property~\ref{prop2}, $(X,Y) \not\models \neg a$.

Now, it remains to prove that any interpretation $(X',Y')$ different from $(X,Y)$ is a model of $\rXY$. Note first that as $Y-X$ is not empty, $\rXY$ is a tautology in classical logic, since its head contains at least a disjunction like $a \vee \neg a$. Therefore, $Y \models \rXY$ and we just have to show that $(X',Y') \models head(\rXY)$ whenever $(X',Y') \models body(\rXY)$. If we assume the latter, from Lemma~\ref{lem:bodyrI} we conclude $X \subseteq X' \subseteq Y' \subseteq Y$, but as $(X',Y')$ is different from $(X,Y)$ then either $X \subset X'$ or $Y' \subset Y$. Assume first that there exists some atom $d \in X'-X$. Since $X' \subseteq Y$, we get $d \in Y-X$, and so $d$ is one of the atoms $a$ in the head of $\rXY$. As $d \in X'$, we have $(X',Y') \models d$ and so $(X',Y') \models head(\rXY)$. On the other hand, if we assume that there exists some $d \in Y-Y'$, as $X \subseteq Y'$, we conclude $d \in Y-X$ and so, $d$ is again one of the atoms in $head(\rXY)$. Now, as $d \not\in Y'$, by Property~\ref{prop2}, $(X',Y') \models \neg d$ and $(X',Y') \models head(\rXY)$.
\end{enumerate}
\end{proof}

A set $S$ of interpretations is {\em total-closed} if given any total $(Y,Y) \in S$ then also $(X,Y)\in S$ for any $X\subseteq Y$.
Clearly, a theory has a total-closed set of countermodels by Property~\ref{prop1}. Then the main theorem immediately follows from the following theorem.

\begin{theorem}\label{th:cm}
Each total-closed set $S$ of interpretations is the set of countermodels of a
nonnested logic program:
\begin{eqnarray*}
\Pi(S) \eqdef \{\rXY \ | \ (X,Y) \in S\}
\end{eqnarray*}
\end{theorem}

\begin{proof}
Any interpretation $(X,Y)$ is a countermodel of $\rXY$ by
Proposition~\ref{prop:cm}, 
so every element of $S$ is a countermodel of $\Pi(S)$. Now take any 
countermodel $(X,Y)$ of $\Pi(S)$. By construction of $\Pi(S)$ and Proposition~\ref{prop:cm} this means that either $\rXY$ or $r_{Y,Y}$ belongs to $\Pi(S)$. Consequently, either
$(X,Y)$ or $(Y,Y) $ is element of $S$. Since $S$ is total-closed, we conclude
that $(X,Y) \in S$.
\end{proof}

For instance, consider formula~(\ref{or-rule}). As its 6 countermodels
are
$$
(\emptyset, \{q\}),\quad
(\{q\}, \{q\}),\quad
(\{q\}, \{p,q\}),\quad
(\emptyset, \{q,r\}),\quad
(\{q\}, \{q,r\}),\quad
(\{q\}, \{p,q,r\}),
$$
it is strongly equivalent to the 6 rules 
\begin{eqnarray*}
\neg p\wedge \neg r &\implies & q\vee\neg q\\
q\wedge \neg p\wedge \neg r &\implies & \bot\\
q\wedge \neg r &\implies & p\vee\neg p\\
\neg p &\implies & q\vee\neg q\vee r\vee \neg r\\
q\wedge \neg p &\implies & r\vee \neg r\\
q &\implies & p\vee\neg p\vee r\vee \neg r.
\end{eqnarray*}

The one-to-one correspondence between total-closed sets of intepretations and the ``classes'' of strongly equivalent programs shown by Theorem~\ref{th:cm} can be used to count the number of such classes.

\begin{theorem}
The number of different logic programs (modulo strong equivalence) that can be built for a finite signature of $n$ atoms is:
\begin{eqnarray*}
\prod^n_{i=0} {\big( 2^{2^i-1} + 1 \big)}^{
\binom{n}{i}
}
\end{eqnarray*}
\end{theorem}
\begin{proof}
We need to count the number of total-closed sets $S$ of interpretations.
For any $S$, let $S_Y$ be the subset of $S$ consisting of the elements of the
form $(X,Y)$. Clearly
each $S_Y$ is independent from $S_{Y'}$ if $Y'\not = Y$, so the number of
values of $S$ is the product of the number of values of $S_Y$ for each
$Y\subseteq \Sigma$.
If $(Y,Y)\not \in S_Y$ then $S_Y$ can independently contain or not each
term of the form $(X,Y)$
where $X$ is a proper subset of $Y$. There are $2^{|Y|}-1$ of such subsets
giving $2^{2^{|Y|}-1}$ total combinations. On the other hand, when $(Y,Y) \in S$, $S_Y$ is completely determined as $\{(X,Y) \ |\ X\subseteq Y\}$, since $S$ is total-closed. Consequently the total number of values for
$S$ is
\begin{eqnarray*}
\prod_{Y\subseteq \Sigma} {\big( 2^{2^{|Y|}-1} + 1 \big)}
\end{eqnarray*}
The final result is obtained by grouping all sets $Y$ with the same
cardinality, and by noticing that the sets $Y$ of size $i$ are $\binom{n}{i}$.
\end{proof}

\section{Normal Forms}
\label{sec:nf}

The reader may have noticed that Theorem~\ref{th:cm} seems, in principle,
stronger than the original claim: we know now that any theory is strongly
equivalent to a {\em nonnested} logic program. Nevertheless, this was also
implicitly asserted by Theorem~\ref{th:gen}, since as we had seen, a nested
program can always be transformed into a nonnested one under strong
equivalence~\cite{Lif99d}. Thus, both proofs actually point out that nonnested
logic programs act as a kind of normal form for the logic of here-and-there. Furthermore,
they show a strong analogy to CNF in classical logic: we can understand the
program as a conjunction of nonnested rules which, in their turn, are seen as
clauses. The method in Section~\ref{sec:nonnested} is then completely analogous
to the construction of the classical CNF of a formula starting from its
countermodels. In both cases, we build a rule/clause per each
countermodel, and this clause refers to all the atoms of the signature. 

In classical logic, however, we know that the construction of CNF from the theory countermodels is completely dual to the construction of DNF from its models. So, the following question arises immediately: is there some kind of {\em disjunctive normal form} for the logic of here-and-there that can be built starting from the theory models? The answer is affirmative, as we show next.

\begin{define}
Given an interpretation $(X,Y)$ for a finite signature $\Sigma$, let
$\cXY$ be the formula
$$
\big(\bigwedge_{a\in X} a\big)\wedge\big(\bigwedge_{b\in \Sigma-Y} \neg b\big)
\wedge \big(\bigwedge_{c\in Y-X} \neg\neg c\big)\wedge 
\bigwedge_{d,e\in Y-X} (d\implies e).
$$\qed
\end{define}

For instance, if $\Sigma=\{p,q,r\}$ then $c_{\{q\}, \{p,q\}}$ is
$$
q\wedge \neg r\wedge \neg \neg p \wedge (p\implies p).
$$

The interesting properties of $\cXY$ are stated in the next Peoposition and
Theorem.
\begin{proposition}\label{lemmac}
The only interpretations that satisfy $\cXY$ are $(X,Y)$ and
$(Y,Y)$.
\end{proposition}
\begin{proof}
Consider the following formulas $c'_{X,Y}$ and $c''_{X,Y}$:
$$
c'_{X,Y}  \eqdef \bigwedge_{c\in Y-X} \neg\neg c\qquad\qquad
c''_{X,Y} \eqdef \bigwedge_{d,e\in Y-X} (d\implies e)
$$
Clearly, $\cXY = body(\rXY)\wedge c'_{X,Y} \wedge c''_{X,Y}$.
We are going to see for which conditions the three terms $body(\rXY)$,
$c'_{X,Y}$ and $c''_{X,Y}$ are satisfied by an interpretation $(X',Y')$. Each
formula $\neg\neg c$ in $c'_{X,Y}$ is satisfied by $(X',Y')$ iff $Y'\models c$
by Property~\ref{prop2}. That means that $(X',Y')\models c'_{X,Y}$ iff
$Y-X\subseteq Y'$, or, equivalently, iff $Y\subseteq Y'\cup X$. On the other
hand, by Lemma~\ref{lem:bodyrI},
$(X',Y')\models body(\rXY)$  iff
$X\subseteq X'\subseteq Y'\subseteq Y$. Notice that if
$X\subseteq Y'$ then $Y'\cup X=Y'$. Consequently,
$(X',Y')\models body(\rXY)\wedge c'_{X,Y}$ iff
$$
X\subseteq X'\subseteq Y'=Y.
$$
That means that the only models for $c_{X,Y}$ are the ones of the form $(X',Y)$
with $X\subseteq X'\subseteq Y$, that satisfy $c''_{X,Y}$. Since $Y$ contains
all the consequents of the implications of $c''_{X,Y}$, then
$Y\models c''_{X,Y}$. If we consider that $X$ doesn't contain any of the
antecedents, then $(X,Y)\models c''_{X,Y}$, and also $(Y,Y)\models c''_{X,Y}$
by Property~\ref{prop1}. It remains to show that if $X\subset X'\subset Y$ then
$(X',Y)\not\models c''_{X,Y}$. This is immediate, since
$c''_{X,Y}$ contains an implication $d\implies e$ with
$d\in X'$ and $e\not\in X'$.
\end{proof}

\begin{theorem}\label{prop-dnf}
Let $T$ be a theory over a finite signature and let $F(T)$ denote the formula
constructed from the models of $T$:
\begin{eqnarray*}
F(T) \eqdef \bigvee_{(X,Y)~:~(X,Y) \models T} \cXY.
\end{eqnarray*}
Then $T$ is equivalent to $F(T)$ under the logic of here-and-there.
\end{theorem}
\begin{proof}
Take any model $(X,Y)$ of $T$. Then, $\cXY$ is a disjunctive term in $F(T)$.
Since $(X,Y)\models \cXY$ by Proposition~\ref{lemmac}, we can conclude that
$(X,Y)\models F(T)$. Now take any model $(X,Y)$ of $F(T)$. That means that,
by Proposition~\ref{lemmac}, $(X,Y)$ satisfies some disjunctive term
$c_{X',Y}$ of $F(T)$, where
\begin{itemize}
\item
$X=X'$, if $X\subset Y$, and
\item
$X'\subseteq Y$, if $X=Y$.
\end{itemize}
We know that $(X',Y)\models T$ by construction of $F(T)$. Consequently,
if $X\subset Y$ then $(X,Y)$ is a model of $T$ because $(X',Y)=(X,Y)$.
For the case $X=Y$, it is sufficient to notice that $(X,Y)=(Y,Y)$, and
that $(X',Y)\models T$ implies $(Y,Y)\models T$ by Property~\ref{prop1}.
\end{proof}

For instance, consider formula~(\ref{or-rule}). As its 21 models are
$$
\begin{array}c
(\emptyset, \emptyset),\quad
(\emptyset, \{p\}),\quad
(\{p\}, \{p\}),\quad
(\emptyset, \{r\}),\quad
(\{r\}, \{r\}),\quad
(\emptyset, \{p,q\}),\quad
(\{p\}, \{p,q\}),\\
(\{p,q\}, \{p,q\}),\quad
(\{p\}, \{q,r\}),\quad
(\{p,q\}, \{q,r\}),\quad
(X, \{p,r\}),\quad
(Y, \{p,q,r\}),\\
\text{($X\subseteq \{p,r\}$ and $Y\subseteq \{p,q,r\}, Y\not = \{q\}$)}
\end{array}
$$
it is strongly equivalent to the disjunction of
$$
\begin{array}c
\neg p\wedge \neg q\wedge \neg r\\
\neg q\wedge \neg r\wedge \neg \neg p\wedge (p\implies p)\\
p\wedge \neg q\wedge \neg r\\
\neg p\wedge \neg q\wedge \neg \neg r\wedge (r\implies r)\\
\neg r\wedge \neg\neg p\wedge \neg \neg q\wedge (p\implies p)\wedge
(p\implies q) \wedge (q\implies p)\wedge (q\implies q)\\
\vdots
\end{array}
$$

As a final remark, note that $F(T)$ is a disjunction of clauses $\cXY$ which are in their turn conjunctions of expressions of a very restricted type: each one may involve negation, implications, and (at most two) atoms, but no conjunction or disjunction. It must be said, however, that we still ignore whether $F(T)$ may have any practical interest at all for computing answer sets.


\section{Conclusions}
\label{sec:conc}

We have shown that when we consider answer sets for arbitrary propositional theories (that is, {\em Equilibrium Logic}) we actually obtain the same expressiveness than when we just deal with disjunctive programs that allow negation in the head. To this aim, we proved that for any propositional theory, there always exists a {\em strongly equivalent} logic program, i.e., a program that may safely replace the original theory in terms of answer sets, even in the context of additional information. In fact, to be precise, our result shows that disjunctive logic programs (with negation in the head) constitute a conjunctive normal form for the monotonic basis of Equilibrium Logic, the intermediate logic of \emph{here-and-there}. 

We actually provided two different proofs for this result: one consisting in a recursive syntactic transformation, and a second one that deals with countermodels of the propositional theory in the logic of {\em here-and-there}. The reason for presenting the two proofs is that, in both cases, they provide technical tools for obtaining additional results. For instance, the proof based on the recursive transformation can be adapted to generalize the Completion Lemma for logic programs from~\cite{Fer05c} to the case of propositional theories: the new statement of the lemma is almost as general as the statement from~\cite{Fer05b}. On the other hand, the technique based on countermodels has helped us to establish for the first time, as far as we know, the number of different programs (modulo strong-equivalence) that can be built with a given number of atoms. Furthermore, this technique has also opened research work under development~\cite{Cab06} that studies the generation of a minimal program from a set of countermodels, analogously to the methods~\cite{Qui52,McC56} for minimizing boolean functions in classical logic, well-known in digital circuit design.

The main focus in this technical note was just to prove the existence of a strongly equivalent program for any propositional theory, leaving more detailed related topics to be treated in subsequent work. Consequently, the methods we present to obtain the resulting program are mostly thought to achieve simple proofs of the main result, rather than to obtain an efficient computation or a more compact representation of the final program itself. As an example of work directly derived from the current result, but perhaps more interesting from a computational viewpoint,~\cite{Cab05} presented a pair of alternative syntactic transformations to reduce a propositional theory into a strongly equivalent logic program. One of the transformations preserves the vocabulary of the original theory, but is exponential in the obtained program size, whereas the other is polynomial, but with the cost of adding auxiliary atoms. \cite{Cab05} further confirms this complexity result for nonnested programs, actually proving that \emph{no polynomial time transformation} to nonnested programs can be obtained if we preserve the original vocabulary. The existence of polynomial space/time strongly equivalent translations \change{to programs} with nested expressions is an open \change{question}.

\section*{Acknowledgements}
Special thanks to Vladimir Lifschitz and David Pearce for their invaluable help to improve the paper; in fact, the motivation for the current research was originated by their joint discussions. The first author is partially supported by the Spanish MEC project TIN-2006-15455-C03-02, and the second by the National Science Foundation under Grant IIS-0412907.

\bibliography{TPLPCabFer07}
\end{document}